\documentclass[conference]{IEEEtran}
\IEEEoverridecommandlockouts

\usepackage{amsmath}
\usepackage{amssymb}
\usepackage{amsthm}
\usepackage{bm}
\usepackage{booktabs}
\usepackage{breqn}
\usepackage[date=year,doi=false,isbn=false,maxbibnames=99,style=ieee,url=false]{biblatex}
\usepackage{graphicx}
\usepackage{multirow}
\usepackage{orcidlink}
\usepackage{subcaption}
\usepackage{hyperref}
\usepackage[capitalize]{cleveref}

\graphicspath{{images/}}

\newcommand{\header}[1]{\multicolumn{1}{c}{#1}}

\makeatletter
\newif\if@anonymize

\@anonymizefalse

\if@anonymize
    \newcommand{\ack}{}
    \newcommand{\authorinfo}{
        \IEEEauthorblockN{Anonymous Author(s)}
        \IEEEauthorblockA{Affiliation\\Address\\\texttt{email}}
    }
    \newcommand{\repourl}{\href{https://github.com/blackblitz/bcl}{\texttt{https://anonymous.4open.science/r/bcl-7693}}}
\else
    \newcommand{\ack}{
        \section*{Acknowledgements}
        
        We thank Pengcheng Hao, Professor Yang Li from Tsinghua Shenzhen International Graduate School and anonymous reviewers for their feedback.
    }
    \newcommand{\authorinfo}{
        \IEEEauthorblockN{Menghao Waiyan William Zhu\orcidlink{0009-0001-4180-5791}}
        \IEEEauthorblockA{Tsinghua Shenzhen International Graduate School\\Shenzhen, China\\\texttt{zhumh22@mails.tsinghua.edu.cn}}
        \and
        \IEEEauthorblockN{Ercan Engin Kuruoğlu}
        \IEEEauthorblockA{Tsinghua Shenzhen International Graduate School\\Shenzhen, China\\\texttt{kuruoglu@sz.tsinghua.edu.cn}}
    }
    \newcommand{\repourl}{\href{https://github.com/blackblitz/bcl}{\texttt{https://github.com/blackblitz/bcl}}}
\fi

\makeatother

\begin{document}

\title{On Sequential Maximum a Posteriori Inference for Continual Learning}

\author{\authorinfo}

\maketitle

\begin{abstract}
We formulate sequential maximum a posteriori inference as a recursion of loss functions and reduce the problem of continual learning to approximating the previous loss function. We then propose two coreset-free methods: autodiff quadratic consolidation, which uses an accurate and full quadratic approximation, and neural consolidation, which uses a neural network approximation. These methods are not scalable with respect to the neural network size, and we study them for classification tasks in combination with a fixed pre-trained feature extractor. We also introduce simple but challenging classical task sequences based on Iris and Wine datasets. We find that neural consolidation performs well in the classical task sequences, where the input dimension is small, while autodiff quadratic consolidation performs consistently well in image task sequences with a fixed pre-trained feature extractor, achieving comparable performance to joint maximum a posteriori training in many cases.
\end{abstract}


\section{Introduction}

When a neural network (including a generalized linear model, which is essentially a neural network with no hidden layers) is trained on a task and fine-tuned on a new task, it loses predictive performance on the old task. This is known as catastrophic forgetting \parencite{mccloskey_catastrophic_1989} and can be prevented by training jointly on all tasks, but the previous data may not be accessible due to computational or privacy constraints. Thus, we would like to learn from a sequence of tasks with limited or no access to the previous data. This is known as continual learning or incremental learning or lifelong learning.

For classification tasks, three types of continual learning settings are commonly studied \parencite{van_de_ven_three_2022}:
\begin{enumerate}
    \item Task-incremental learning, in which task IDs are provided and the classes change between tasks
    \item Domain-incremental learning, in which task IDs are not provided and the classes remain the same between tasks but the input data distribution changes between tasks
    \item Class-incremental learning, in which task IDs are not provided and the classes change between tasks
\end{enumerate}
For example, Split MNIST is a sequence of five tasks created from the MNIST dataset, in which the first task consists of images of zeros and ones, the second task consists of images of twos and threes and so on. In the task-incremental setting, the task IDs 1-5 are provided, and the neural network only has to decide between two classes for each task ID. Typically, a multi-headed neural network with five heads (one head for each task) is used in this setting. In the domain-incremental setting, the task is classification of even and odd digits without access to the task ID. In the class-incremental setting, the task is classification of all ten digits without access to the task ID. In these settings, a single-headed neural network is typically used.

Task-incremental learning has been criticized as task IDs make the problem of continual learning easier \parencite{farquhar_towards_2019}. In fact, if there is only one class per task, then the task ID could be used to make a perfect prediction. Moreover, in practice, it is unlikely that task IDs are accessible. For example, in Split MNIST, we would like to classify all ten digits in the end rather than just tell which of the two digits in each task. In accordance with the desiderata proposed in \cite{farquhar_towards_2019}, we focus on domain- and class-incremental learning with single-headed neural networks on task sequences with more than two similar tasks and no access to the previous data.

For a single-headed neural network with fixed architecture, the learning problem can be formulated as Bayesian inference on the neural network parameters. Then, sequential Bayesian inference provides an eltextegant approach to continual learning. In particular, continual learning can be done by using the previous posterior distribution as the current prior distribution. If full Bayesian prediction is used, then the approach is known as a Bayesian neural network. In our work, we focus on maximum a posteriori (MAP) prediction, which uses only the maximum point of the posterior distribution. By defining the loss function as the negative log joint probability density function (PDF), this leads to a recursive formulation of loss functions, and the problem is reduced to approximating the previous loss function.

When adult humans learn to recognize new objects, they have already learned similar objects before, and they do not need to continually learn low-level features such as edges, corners and shapes. This suggests that the low-level layers of a neural network should be fixed after learning on a similar task, and then the neural network can be used as a feature extractor. For example, a neural network pre-trained on handwritten letters can be used as a feature extractor for continual learning on handwritten digits.

Our goals are to investigate the continual learning performance of a full quadratic approximation and a neural network approximation of the previous loss functions and the effect of using a pre-trained feature extractor for sequential MAP inference.

\section{Sequential Maximum a Posteriori Inference}

We describe our probabilistic model for continual learning and how sequential MAP inference based on it leads to a recursion of loss functions. Then, we propose two methods for approximating the previous loss function: autodiff quadratic consolidation (AQC), which uses an accurate and full quadratic approximation, and neural consolidation (NC), which uses a neural network approximation. Finally, we discuss their limitations.

\subsection{Probabilistic Model}

\begin{figure}[ht]
    \centering
    \begin{tikzpicture}[main/.style={circle, draw, minimum size=10mm}, node distance=15mm]
        \node[main] (theta) {\(\bm\theta\)};
        \node[main] (y1) [below of=theta] {\(\bm y_1\)};
        \node[main] (x1) [below of=y1] {\(\bm x_1\)};
        \node[main] (y2) [right of=y1] {\(\bm y_2\)};
        \node[main] (x2) [right of=x1] {\(\bm x_2\)};
        \node (ydots) [right of=y2] {\(\ldots\)};
        \node (xdots) [right of=x2] {\(\ldots\)};
        \node[main] (yt) [right of=ydots] {\(\bm y_t\)};
        \node[main] (xt) [right of=xdots] {\(\bm x_t\)};
        \draw[->] (theta) -- (y1);
        \draw[->] (theta) -- (y2);
        \draw[->] (theta) -- (yt);
        \draw[->] (x1) -- (y1);
        \draw[->] (x2) -- (y2);
        \draw[->] (xt) -- (yt);
    \end{tikzpicture}
    \caption{Bayesian network for continual learning. \(\bm\theta\) is the collection of parameters of the neural network, \(\bm x_{1:t}\) are the inputs and \(\bm y_{1:t}\) are the outputs.}
    \label{fig:probmodel}
\end{figure}

Let \(\bm\theta\) be the collection of parameters of the neural network and \(\bm x_{1:t}\) and \(\bm y_{1:t}\) be the inputs and outputs from time \(1\) to \(t\), respectively. \(\bm x_{1:t}\) are assumed to be independent, and \(\bm y_{1:t}\) are assumed to be conditionally independent given \(\bm\theta\) and \(\bm x_{1:t}\). These assumptions are depicted in \cref{fig:probmodel}.

\((\bm x,\bm y)_{1:t}\) are not necessarily identically distributed. However, it is assumed that tasks are similar, i.e. the form of the likelihood function \(l_t(y_t|\theta,x_t)\) is the same for all tasks. For example, in multi-class classification, it is the categorical likelihood function for all tasks.

After observing \((\bm x,\bm y)_{1:t}=(x,y)_{1:t}\) at time \(t\), the posterior PDF is
\begin{equation}
    p_t(\theta|x_{1:t},y_{1:t})=\frac1{z_t}p_{t-1}(\theta|x_{1:t-1},y_{1:t-1})l_t(y_t|\theta,x_t)\label{eq:bayes}
\end{equation}
where \(z_t=\int_{\Theta}p_{t-1}(\theta|x_{1:t-1},y_{1:t-1})l_t(y_t|\theta,x_t)d\theta\) is a normalization term which does not depend on \(\theta\). MAP prediction uses the maximum of the posterior PDF to make a prediction \(f(x;\theta^*_t)\), where \(f\) is the neural network function, \(x\) is the input and \(\theta^*_t\) is the MAP estimate at time \(t\).

Since multiplying by a constant does not affect the minimum, the loss function \(\mathfrak L_t\) at time \(t\) can be defined as the negative log joint PDF \(-\ln j_t(\theta,y_{1:t}|x_{1:t})\) at time \(t\). Then, the minimum of the loss function is equivalent to the MAP estimate, and we have a recursion of loss functions for \(t=1,2,\ldots\):
\begin{equation}
    \mathfrak L_t(\theta)=\mathfrak L_{t-1}(\theta)+\mathfrak l_t(\theta)\label{eq:lossrecursion}
\end{equation}
where \(\mathfrak l_t\) is the negative log likelihood (NLL) at time \(t\). For binary classification, where the likelihood is assumed to be Bernoulli, \(\mathfrak l_t\) is the binary or Bernoulli cross entropy, while for multi-class classification, where the likelihood is assumed to be categorical. \(\mathfrak l_t\) is the categorical cross entropy. \(\mathfrak L_0\) is the negative log (un-normalized) prior at time \(1\), e.g., \(\frac12\Vert\theta\Vert^2\) for the standard Gaussian prior. 

In \cref{eq:lossrecursion}, \(\mathfrak L_{t-1}\) depends on the previous data \((\bm x,\bm y)_{1:t-1}\) and \(\mathfrak l_t\) depends on the current data \((\bm x,\bm y)_t\). Forgetting happens when we minimize only \(\mathfrak l_t\) but ignore \(\mathfrak L_{t-1}\) as in fine-tuning. In joint MAP training, all the data are used, so \(\mathfrak L_t\) is effectively minimized. If there is no access to the previous data, then \(\mathfrak L_{t-1}\) must be approximated. We investigate two ways to approximate it, namely quadratic approximation and neural network approximation.

\subsection{Autodiff Quadratic Consolidation}
\label{subsec:aqc}

A quadratic approximation of \(\mathfrak L_{t-1}\) corresponds to a Laplace approximation of the posterior PDF at time \(t-1\). The quadratic approximation is a second-order Taylor series approximation around \(\theta^*_{t-1}\), where the gradient is zero, so the linear term disappears. Moreover, the constant term does not affect the PDF. Thus, the approximation consists of a single quadratic term of the form \(\frac12(\theta-\theta^*_{t-1})^TH(\mathfrak l_{t-1})(\theta^*_{t-1})(\theta-\theta^*_{t-1})\), where \(H(\mathfrak l_{t-1})(\theta^*_{t-1})\) is the Hessian matrix of the NLL of task \(t-1\) at \(\theta^*_{t-1}\). \cite{huszar_quadratic_2017} shows that successive quadratic approximation results in addition of the Hessian matrices of the NLL of the previous tasks at the corresponding minima. Thus, the approximate loss function is
\begin{equation}
    \hat{\mathfrak L}_t(\theta)=\frac\lambda2(\theta-\theta^*_{t-1})^T\left(\sum_{i=0}^{t-1}H(\mathfrak l_i)(\theta^*_i)\right)(\theta-\theta^*_{t-1})+\mathfrak l_t(\theta)\label{eq:aqc}
\end{equation}
where \(\lambda\) is a positive real number introduced as a hyperparameter.

The Hessian matrix is the transpose of the Jacobian matrix of the gradient: \(H(f)(x)=(J(\nabla f)(x))^T\) for any twice differentiable point \(x\) of a function \(f\). In most cases, the NLL \(\mathfrak l_{t-1}\) is twice continuously differentiable, so the Hessian matrix at its minimum is symmetric positive definite. Then, it can be implemented as \(J(\nabla\mathfrak l_{t-1})(\theta^*_{t-1})\), the Jacobian matrix of its gradient at its minimum, by using automatic differentiation.

Since the Hessian operator is linear, the Hessian matrix of the batch negative log likelihood is equal to the sum of the Hessian matrices of the mini-batch negative log likelihood:
\begin{equation}
    H(\mathfrak l_{t-1})(\theta^*_{t-1})=H\left(\sum_{j=1}^b\mathfrak l_{t-1,j}\right)(\theta^*_{t-1})=\sum_{j=1}^bH(\mathfrak l_{t-1,j})(\theta^*_{t-1})\label{eq:hessian}
\end{equation}
where \(\mathfrak l_{t-1,j}\) is the mini-batch negative log likelihood of the \(j\)-th mini-batch at time \(t-1\).

The above trick allows working with a large dataset by dividing it into small mini-batches. However, if the neural network is large, i.e. \(\bm\theta\) has a large amount of parameters, then the computation may become intractable.

The training for each task thus consists of three steps:
\begin{enumerate}
    \item If it is the first task, then the loss function is set to \(\frac12\lVert\theta\rVert^2+\mathfrak l_1(\theta)\) assuming a standard Gaussian prior; otherwise, it is updated as in \cref{eq:aqc} with the MAP estimate \(\theta^*_{t-1}\) and the Hessian matrix \(H_{t-1}=\sum_{i=0}^{t-1}H(\mathfrak l_i)(\theta^*_i)\) of the previous task.
    \item Training is done on the loss function using mini-batch gradient descent, and the regularization term is scaled by dividing by the number of mini-batches in the dataset.
    \item The Hessian matrix for the current task is computed and added to that of the previous task \(H_t=H_{t-1}+H(\mathfrak l_t)(\theta^*_t)\), and the current MAP estimate \(\theta^*_t\) and Hessian matrix \(H_t\) are stored in order to be used to update the next loss function.
\end{enumerate}

This method is referred to as Autodiff Quadratic Consolidation (AQC).

\subsection{Neural Consolidation}
\label{subsec:nc}

A neural network approximation uses a consolidator neural network \(\kappa\) with parameters \(\phi^*_{t-1}\). The approximate loss function is
\begin{equation}
    \hat{\mathfrak L}_t(\theta)=\lambda\kappa(\theta;\phi^*_{t-1})+\mathfrak l_t(\theta)\label{eq:nc}
\end{equation}
where \(\lambda\) is a positive real number introduced as a hyperparameter and \(\phi^*_{t-1}\) is the collection of trained parameters of the consolidator neural network at time \(t-1\).

The consolidator neural network is trained by minimizing an \(L^2\)-regularized Huber loss function to fit it to the previous loss function with a sample generated uniformly within a ball of radius \(r\) around \(\theta^*_{t-1}\) at each gradient descent step. If \(n\) points are generated, then the consolidator loss function is
\begin{equation}
    \mathfrak L_{t-1,\kappa}(\phi)=\frac12\beta\lVert\phi\rVert_2^2+\sum_{i=1}^nh_{t-1,i}(\phi)
    \label{eq:consolidator}
\end{equation}
where \(\beta\) is a positive real number introduced as a hyperparameter and \(h_{t-1,i}(\phi)\) is the Huber loss function with respect to \(\hat{\mathfrak L}_{t-1}(\theta_i)\) and \(\kappa(\theta_i;\phi)\). If the dataset is very large, \(\hat{\mathfrak L}_{t-1}\) can be computed on the sample in mini-batches and added.

The training for each task thus consists of three steps:
\begin{enumerate}
    \item If it is the first task, then the loss function is set to \(\frac12\lVert\theta\rVert^2+\mathfrak l_1(\theta)\) assuming a standard Gaussian prior; otherwise, it is updated as in \cref{eq:nc} with the parameters \(\phi^*_{t-1}\) of the consolidator neural network of the previous task.
    \item Training is done on the loss function using mini-batch gradient descent, and the regularization term is scaled by dividing by the number of mini-batches in the dataset.
    \item The consolidator neural network is trained by performing gradient descent on \cref{eq:consolidator} with a sample of \(n\) points of \(\theta\) generated uniformly within a ball of radius \(r\) around \(\theta^*_t\) at each step as described above, and the parameters \(\phi^*_t\) of the consolidator neural network are stored in order to be used to update the next loss function.
\end{enumerate}

This method is referred to as Neural Consolidation (NC).

\subsection{Limitations}

As in all single-headed approaches, the total number of classes must be known in advance. The main limitation of AQC and NC is that they are not scalable with respect to the neural network size although they can be used when the datasets are large. This limitation may be overcome by using a fixed feature extractor pre-trained on a similar task and performing continual learning with one dense layer on the features. Finally, both continual learning methods are sensitive to hyperparameters, so a validation dataset sequence should be used to perform hyperparameter tuning.

\section{Related Work}
\label{sec:rw}

There are several continual learning methods which are based on sequential MAP inference and use quadratic approximation of the previous loss function. Elastic weight consolidation (EWC) approximates the Hessian matrix by using a diagonal approximation of the empirical Fisher information matrix (eFIM) \parencite{kirkpatrick_overcoming_2017}. The original EWC adds a quadratic term to the objective for every task. \cite{huszar_note_2018} proposes a corrected objective with a single quadratic term for which the eFIM can be cumulatively added. There is another variant called EWC++ \parencite{chaudhry_riemannian_2018}, which performs a convex combination of the previous and current eFIMs instead of adding them. Synaptic intelligence (SI) performs a diagonal approximation of the Hessian matrix by using the change in loss during gradient descent \parencite{zenke_continual_2017}. Online structured Laplace approximation uses Kronecker factorization to perform a block-diagonal approximation of the Hessian matrix, in which the diagonal blocks of the matrix correspond to a layer of the neural network \parencite{ritter_online_2018}.

Another class of methods that are not based on sequential MAP inference but are based on sequential Bayesian inference is sequential variational inference. It approximates the posterior distribution with a variational distribution, which is a simple parametric distribution, typically a Gaussian or a Gaussian mixture distribution, by minimizing an objective called the variational free energy or the negative evidence lower bound with respect to the parameters of the variational distribution. It uses the whole posterior distribution rather than a point from it to make predictions. Gaussian variational continual learning (G-VCL) \parencite{nguyen_variational_2018} and Gaussian mixture variational continual learning (GM-VCL) \parencite{phan_reducing_2022} approximate the posterior distribution over the parameters with a Gaussian distribution and a Gaussian mixture distribution, respectively. Gaussian sequential function space variational inference (G-SFSVI) \parencite{rudner_continual_2022} approximates the posterior distribution over the outputs (before the final activation function) of a selected number of inputs called inducing points with a Gaussian distribution.

Pre-training for initialization and pre-training for feature extraction have both been empirically shown to improve continual learning. In the former, the pre-trained parameters are used as the initial parameters for continual learning \parencite{lee_pre-trained_2023,mehta_empirical_2023}. In the latter, the pre-trained neural network is used as a feature extractor \parencite{hu_continual_2021,li_continual_2022,yang_continual_2023}.

We investigate the continual learning performance of AQC, which is the most accurate form of quadratic approximation of the previous loss function, and NC, which is a neural network approximation of the previous loss function. In image classification tasks, we use a fixed feature extractor pre-trained on a similar task and perform continual learning with one dense layer on the features.

\section{Experiments}

\begin{figure*}[ht]
    \begin{subfigure}{\textwidth}
        \includegraphics[width=\textwidth]{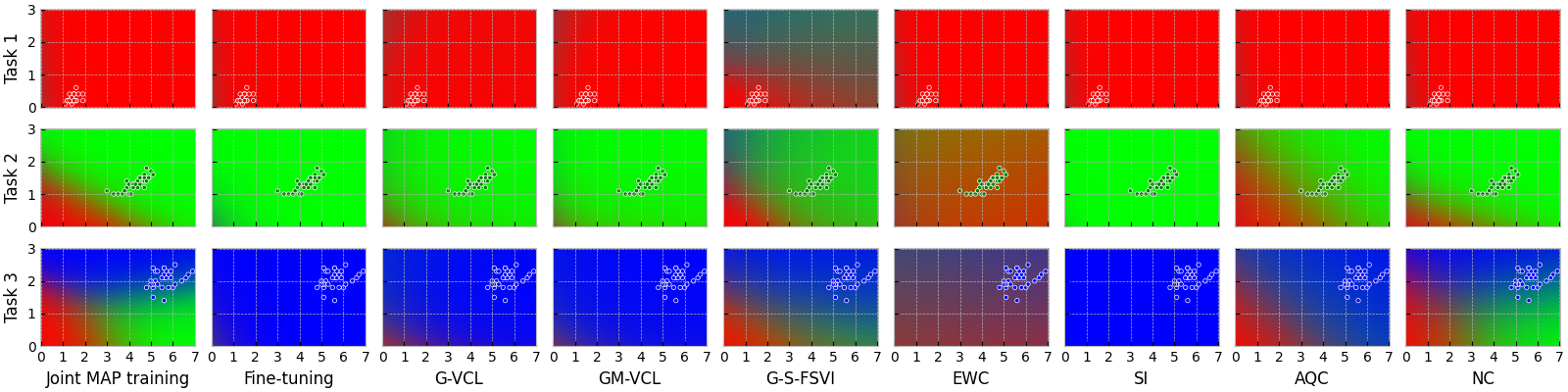}
        \caption{Softmax regression}
        \centering
    \end{subfigure}
    \begin{subfigure}{\textwidth}
        \includegraphics[width=\textwidth]{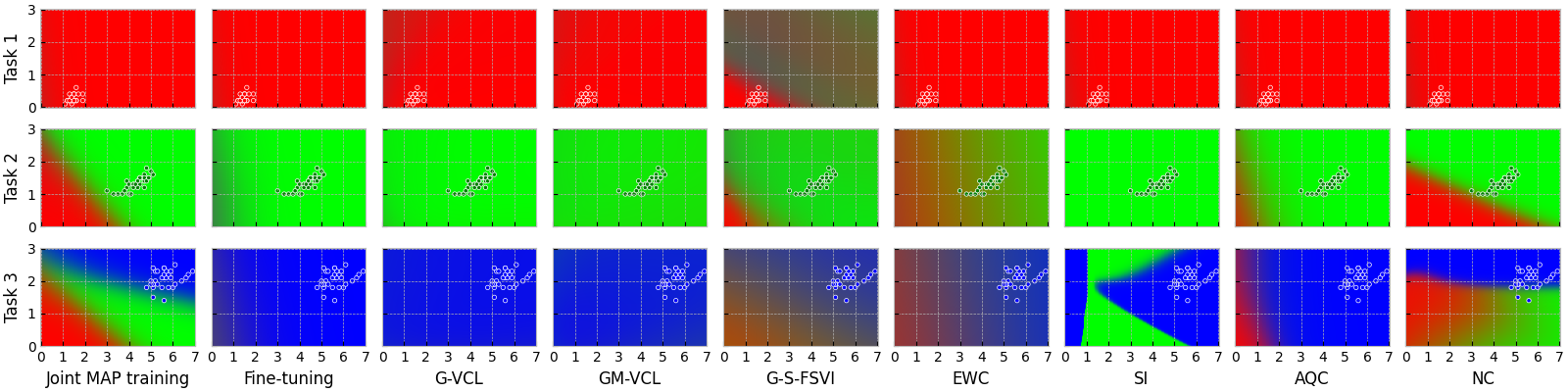}
        \caption{Fully connected neural network}
        \centering
    \end{subfigure}
    \caption{Visualizations of prediction probabilities for the methods on CI Split 2D Iris. The x-axis is the petal length (cm) and the y-axis is the petal width (cm). The pseudocolor plot shows the prediction probabilities, where the 3 class probabilities are mapped to the red, green and blue values, respectively, and the dots show the observed data. NC performs the best and is better with softmax regression.}
    \label{fig:viz}
\end{figure*}

\begin{table*}[ht]
    \caption{Testing final average accuracy for the methods on classical and image task sequences. For classical task sequences, results for softmax regression (SR) and a fully connected neural network (FCNN) are shown. Pre-training on a similar task is used for image task sequences. NC performs the best in most classical task sequences while AQC performs the best in all image task sequences with a fixed pre-trained feature extractor.}
    \label{tab:faa}
    \begin{subtable}{\textwidth}
        \caption{Classical task sequences}
        \centering
        \begin{tabular}{lrrrrrrr}
        \toprule
        \header{\multirow{2}{*}{Method}} & \multicolumn{2}{c}{CI Split Iris} & \multicolumn{2}{c}{CI Split Wine}\\
        & \header{SR} & \header{FCNN} & \header{SR} & \header{FCNN}\\
        \midrule
        Joint MAP training & 96.6667 & 100.0000 & 91.1111 & 91.1111\\
        Fine-tuning & 33.3333 & 33.3333 & 33.3333 & 33.3333\\
        \midrule
        G-VCL & 33.3333 & 33.3333 & 33.3333 & 33.3333\\
        GM-VCL & 33.3333 & 33.3333 & 33.3333 & 33.3333\\
        G-SFSVI & 66.6667 & 33.3333 & 33.3333 & 33.3333\\
        \midrule
        EWC & 63.3333 & 33.3333 & 46.6667 & 33.3333\\
        SI & 33.3333 & 33.3333 & 33.3333 & 33.3333\\
        AQC & 66.6667 & \textbf{63.3333} & 49.6032 & 33.3333\\
        NC & \textbf{93.3333} & \textbf{63.3333} & \textbf{62.6984} & \textbf{48.2540}\\
        \bottomrule
        \end{tabular}
    \end{subtable}
    \newline
    \vspace*{0.2cm}
    \newline
    \begin{subtable}{\textwidth}
        \caption{Image task sequences}
        \centering
        \begin{tabular}{lrrrrrrrrrr}
        \toprule
        \header{Method} & \header{CI Split MNIST} & \header{CI Split CIFAR-10} & \header{CI Split HAM-8} & \header{DI Split MNIST} & \header{DI Split CIFAR-8} & \header{DI Split HAM-6}\\
        \midrule
        Joint MAP training & 95.1077 & 76.2500 & 43.0531 & 93.4338 & 96.6250 & 68.3237\\
        Fine-tuning & 19.8382 & 19.0400 & 21.9512 & 64.4789 & 89.6250 & 63.8068\\
        \midrule
        G-VCL & 25.8657 & 19.0100 & 21.6463 & 80.1092 & 92.9000 & 63.3071\\
        GM-VCL & 25.2141 & 18.9800 & 21.0366 & 77.3996 & 92.6000 & 63.6112\\
        G-SFSVI & 51.1918 & 32.8000 & 34.6563 & 72.5471 & 91.2375 & 63.3381\\
        \midrule
        EWC & 73.3744 & 31.9400 & 30.8636 & 82.9540 & 94.6625 & 63.8068\\
        SI & 30.2364 & 27.0500 & 21.6463 & 79.4063 & 95.0750 & 64.0020\\
        AQC & \textbf{92.5394} & \textbf{61.3100} & \textbf{41.6660} & \textbf{90.9951} & \textbf{96.2000} & \textbf{66.1442}\\
        NC & 77.5334 & 40.2500 & 24.1040 & 89.9168 & 93.8000 & 62.2400\\
        \bottomrule
        \end{tabular}
    \end{subtable}
\end{table*}

Experiments are performed on classical task sequences as well as image task sequences. In each experiment, the final average accuracy of AQC and NC are compared with reference methods, sequential variational inference methods and sequential MAP inference methods. Each task sequence has a training dataset sequence, validation dataset sequence and testing dataset sequence. The validation dataset sequence is used to perform hyperparameter tuning via grid search. Descriptions of data used and methods compared as well as adiscussion of results are provided below, and more details are provided in \cref{sec:exp}.

\subsection{Data}

The task sequences for continual learning as well as the tasks for pre-training are listed below. The classical task sequences that we introduce here might seem simple, but they are challenging task sequences for continual learning. In all task sequences, CI indicates that it is for class-incremental learning, while DI indicates that it is for domain-incremental learning.

\begin{itemize}
    \item Classical Task Sequences
    \begin{itemize}
        \item\textbf{CI Split Iris}: Iris is a task for classification of 3 species of flowers based on 4 features. It is split into 3 tasks for learning 1 species at a time. CI Split 2D Iris is a task sequence for visualization derived from it by selecting two features "petal length" and "petal width".
        \item\textbf{CI Split Wine}: Wine is a task for classification of 3 classes of wine based on 13 features. It is split into 3 tasks for learning 1 class at a time.
    \end{itemize}
    \item{Image task sequences}
    \begin{itemize}
        \item\textbf{CI Split MNIST}: MNIST is a task for classification of 10 classes of digits based on grayscale images of handwritten digits. It is split into 5 tasks for learning 2 classes at a time.
        \item\textbf{CI Split CIFAR-10}: CIFAR-10 is a task for classification of 10 classes of natural objects based on images. It is split into 5 tasks for learning 2 classes at a time.
        \item\textbf{CI Split HAM-8}: HAM10000 is a task for classification of 8 classes of skin conditions based on dermatoscopic images. It is renamed to HAM-8 based on the number of classes and split into 4 tasks for learning 2 classes at a time.
        \item\textbf{DI Split MNIST}: MNIST is split into 5 tasks with 2 classes at a time, but each task is of binary classification of even and odd digits.
        \item\textbf{DI Split CIFAR-8}: CIFAR-10 has 6 types of animal and 4 types of vehicle, so 2 types of animal ("bird" and "frog") are removed to make CIFAR-8, which is then split into 4 tasks with 2 classes at a time, but each task is of binary classification of vehicles and animals.
        \item\textbf{DI Split HAM-6}: HAM-8 has 4 types of benign skin condition, 1 type of indeterminate skin condition and 3 types of malignant skin condition, so 1 type of benign skin condition ("vascular lesion") and 1 type of indeterminate skin condition ("actinic keratosis") are removed to make HAM-6, which is then split into 3 tasks with 2 classes at a time, but each task is of binary classification of benign and malignant skin conditions.
    \end{itemize}
    \item{Pre-training Tasks}
    \begin{itemize}
        \item\textbf{EMNIST Letters}: EMNIST Letters is a task for classification of 26 classes of letters based on grayscale images of handwritten letters. It has no classes in common with MNIST and is used for pre-training for CI Split MNIST and DI Split MNIST.
        \item\textbf{CIFAR-100}: CIFAR-100 is a task for classification of 100 classes of natural objects based on images. It has no classes in common with CIFAR-10 and is used for pre-training for CI Split CIFAR-10 and DI Split CIFAR-10.
        \item\textbf{BCN-12}: BCN20000 is a task for classification of 8 classes of skin conditions based on dermatoscopic images. It is renamed to BCN-12 based on the number of classes. Most classes are in common with HAM-8, but the images are from different populations. It is used for pre-training for CI Split MNIST and DI Split MNIST.
    \end{itemize}
\end{itemize}

\subsection{Methods}

We compare AQC and NC with reference methods, sequential variational inference methods and sequential MAP inference methods. Joint MAP training and fine-tuning serve as the best and worst reference methods, respectively. In the former, all the previous data are used together with the current data to train the neural network, while in the latter, the previously trained neural network is simply fine-tuned to the current task. The variational inference methods compared are G-VCL \parencite{nguyen_variational_2018}, GM-VCL \parencite{phan_reducing_2022} and G-SFSVI \parencite{rudner_continual_2022}, and the sequential MAP inference methods compared are EWC with Huszár's corrected penalty \parencite{huszar_quadratic_2017,huszar_note_2018} and SI \parencite{zenke_continual_2017}. These methods are described in \cref{sec:rw}. To make fair comparisons, only coreset-free methods, i.e. methods do not store any previous data, are considered.

Each method runs through each training dataset sequence, its hyperparameters are selected based on the final average accuracy on the validation dataset sequence, and it is evaluated based on the final average accuracy on the testing dataset sequence. The final average accuracy on a dataset sequence is defined as the average of all the accuracies on all the datasets in the sequence.

\subsection{Results}

Since AQC relies on the very accurate automatic differentiation for Hessian computation, we expect that it has better final average accuracy than EWC and SI, and we are interested in observing how much better it performs. Since neural networks are powerful function approximators, and the previous loss functions in our experiments are not quadratic, we expect that NC has better final average than the quadratic approximation methods. We are also interested in how much a pre-trained feature extractor helps in sequential MAP inference.

Visualizations of the prediction probabilities for the methods on CI Split 2D Iris for softmax regression and a fully connected neural network are shown in \cref{fig:viz}. We find that AQC performs better than EWC and SI, and NC performs the best, but it does better for softmax regression.

The testing final average accuracies for the methods on classical and image task sequences are shown in \cref{tab:faa}. We find that AQC performs better than EWC and SI, and NC performs the best in most classical task sequences. We also find that NC performs better with softmax regression than with a fully connected neural network probably because the loss function in the former is convex and is easier to fit. In image task sequences, where a feature extractor is used, AQC consistently performs the best and has performance comparable to joint MAP training in some task sequences. However, we find that NC does not perform as well as AQC, but is better than EWC and SI in some task sequences.

It is notable that EWC performs as poorly as fine-tuning in class-incremental learning on the whole neural network from scratch \parencite{van_de_ven_three_2022}, but using a pre-trained feature extractor significantly improves it. Moreover, in DI Split CIFAR-8, using a pre-trained feature extractor alone significantly reduces forgetting, and even fine-tuning performs quite well.

A possible reason that NC does not perform well in image task sequences is that the dimension of the feature space is high (64 in CI Split MNIST and DI Split MNIST and 512 in other task sequences), and random sampling becomes inefficient in high dimensions.

\subsection{Data Availability}

All the datasets used in this work are publicly available. Iris and Wine are available from the \texttt{scikit-learn} package \parencite{pedregosa_scikit-learn_2011}, which is released under the 3-clause BSD license. MNIST, EMNIST, CIFAR-10 and CIFAR-100 are available from the \texttt{pytorch} package \parencite{ansel_pytorch_2024}, which is also released under the 3-clause BSD license. HAM10000 \parencite{tschandl_ham10000_2018} is released by the Hospital Clinic in Barcelona under CC BY-NC, and BCN20000 \parencite{hernandez-perez_bcn20000_2024} is released by ViDIR Group, Department of Dermatology, Medical University of Vienna, also under CC BY-NC.

\subsection{Code Availability}

Documented and reproducible code is available under an MIT Licence at \repourl.

\section{Conclusion}

We formulated continual learning based on sequential maximum a posteriori inference as a recursion of loss functions and reduced the problem to function approximation. We then proposed two coreset-free methods based on it: autodiff quadratic consolidation and neural consolidation, which use a full quadratic approximation and a neural network approximation, respectively, to approximate the previous loss function. Moreover, we showed empirically that pre-training the neural network on a similar task could significantly reduce forgetting with sequential maximum a posteriori inference methods. Neural consolidation performs the best in classical task sequences, where the input dimension is small. Autodiff quadratic consolidation consistently performs very well in image task sequences with pre-training on a similar task, achieving performance comparable to joint maximum a posteriori training in many cases. In the future, we may consider special neural network architectures for neural consolidation as well as more applications in medical image classification, document image understanding \parencite{kuruoglu_using_2010} and materials science \parencite{saffarimiandoab_insights_2021}.

\ack

\printbibliography

\clearpage

\appendices

\crefalias{section}{appendix}

\section{Experiment Details}
\label{sec:exp}

\subsection{Data Preparation}

For task sequences based on Iris and Wine, the dataset is split into training and testing datasets with 20\% testing size, and then the training dataset into training and validation datasets with 20\% validation size, so the training, validation and testing proportions are 64\%, 16\% and 20\%, respectively. Finally, each dataset is split by class into a dataset sequence.

For EMNIST Letters, CIFAR-100 and task sequences based on MNIST and CIFAR-10, training and testing datasets are available from PyTorch, so the training dataset is split into training and validation datasets with 20\% validation size. Each dataset is then split by class into a dataset sequence.

For BCN-12 and HAM-8, the \(640\times450\) images are resized to \(32\times32\) with Lanczos interpolation. For all image data, the pixel values are divided by 255 so that they take values between 0 and 1. Data augmentation (e.g. flipping and cropping) is not performed.

\subsection{Neural Network Architectures}

The fully connected neural network used for CI Split 2D Iris and CI Split Iris has 1 hidden layer of 4 nodes, while that used for CI Split Wine has 1 hidden layer of 16 nodes. All the hidden nodes use swish activation.

The pre-trained neural network for both CI Split MNIST and DI Split MNIST has 2 convolutional layers and 2 dense layers, totaling 4 layers. Each convolutional layer has 32 \(3\times3\) filters and is followed by group normalization with 32 groups, swish activation and average pooling with a size of \(2\times2\). The hidden dense layer has 64 nodes with swish activation. Thus, the feature dimension is 64.

The pre-trained neural network for CI Split CIFAR-10, CI Split HAM-8 DI Split CIFAR-8 and DI Split HAM-6 has 17 convolutional layers and 1 dense layer, totaling 18 layers. Each convolutional layer is followed by group normalization with 32 groups and swish activation. The 2nd to the 17th convolutional layers are arranged into 8 residual blocks, each with 2 convolutional layers, and every 2 residual blocks are followed by average pooling with a size of \(2\times2\). The numbers of filters for the 17 convolutional layers are 32, 64, 64, 64, 64, 128, 128, 128, 128, 256, 256, 256, 256, 512, 512, 512 and 512, respectively, and the filter sizes are all \(3\times3\). Thus, the feature dimension is 512.

In all experiments, the consolidator neural network used in NC is a fully connected neural network with 2 hidden layers, each with 256 nodes. All the hidden nodes use swish activation.

\subsection{Training}

In all experiments, the prior PDF at time 1 is a standard Gaussian PDF (of an appropriate dimension), and an Adam optimizer is used with a one-cycle learning rate schedule. The neural network parameters are initialized by using the Lecun normal initializer for the weights and setting to zero for the biases. For pre-training tasks and class-incremental task sequences, each task is of multi-class classification, so categorical cross entropy is used, while for domain-incremental task sequences, each task is of binary classification, so binary or Bernoulli cross entropy is used. BCN-12 is a task with severe class imbalance, so for pre-training on BCN-12, instead of the standard cross entropy, a weighted cross entropy  \(-\sum_{i=1}^k\frac m{n_i}p_i\ln q_i\) is used, where \(p_i\) is the true label indicator and \(q_i\) is the predicted probability, \(n_i\) is the frequency of the \(i\)-th class and \(m=\min\{n_1,n_2,\ldots,n_k\}\).

For CI Split 2D Iris and CI Split Iris, training for each task is performed for 100 epochs with a base learning rate of 0.1 and a batch size of 16. For CI Split Wine, training for each task is performed similarly but with a base learning rate of 0.01.

For CI Split MNIST and DI Split MNIST, pre-training is performed on EMNIST Letters for 20 epochs with a base learning rate of 0.01 and a batch size of 64, and training for each task is performed similarly. For CI Split MNIST and DI Split CIFAR-8, pre-training is performed on CIFAR-100 for 20 epochs with a base learning rate of 0.001 and a batch size of 64, and training for each task is performed similarly but with a base learning rate of 0.01. For CI Split HAM-8 and DI Split HAM-6, pre-training is performed on BCN-12 for 20 epochs with a base learning rate of 0.0001 and a batch size of 64, and training for each task is performed similarly but with a base learning rate of 0.001.

For G-SFSVI, the inducing points are randomly generated from a uniform distribution in a hyperrectangle the boundaries of which are determined by the minimum and maximum values of the training input data across all tasks in the task sequence. For image task sequences with pre-training, the boundaries for each feature component are set to -1 and 6.

\subsection{Hyperparameter Tuning}

In EWC, SI, AQC and NC, there is a hyperparameter \(\lambda\) that determines the regularization strength. SI has an extra damping hyperparameter \(\xi\) and NC has an extra radius hyperparameter \(r\). Hyperparameter tuning is performed based on the validation final average accuracy via grid search among the following values:

\begin{itemize}
    \item EWC: \(\lambda\in\{1,10,100,1000,10000\}\)
    \item SI: \(\lambda\in\{1,10,100,1000,10000\},\xi\in\{0.1,1.0,10\}\)
    \item AQC: \(\lambda\in\{1,10,100,1000,10000\}\)
    \item NC: \(\lambda\in\{1,10,100,1000,10000\},r\in\{1,10,100\}\)
\end{itemize}

\end{document}